\documentclass[10pt, a4paper]{article}

\usepackage[final]{lrec-coling2024} 

\usepackage{graphicx}
\usepackage{tabularx}
\usepackage{soul}
\usepackage{newfloat,caption,float}
\DeclareFloatingEnvironment[
  fileext = prompt,
  listname = Prompt,
  name = Prompt,
  placement = H,
  within = none,
  ]{prompt}
\usepackage{tcolorbox}
\usepackage[export]{adjustbox}
\usepackage{multirow}
\usepackage{booktabs}
\usepackage{inconsolata}
\usepackage{amsmath}
\usepackage{amssymb}
\usepackage[printonlyused]{acronym}
\usepackage{makecell, array}

\usepackage{xcolor}
\usepackage{soul}

\usepackage{hyperref}
 \definecolor{darkblue}{rgb}{0, 0, 0.5}
  \hypersetup{colorlinks=true, citecolor=darkblue, linkcolor=darkblue, urlcolor=darkblue}

\usepackage{xstring}

\usepackage{color}

\usepackage{titlesec}

\title{A Hybrid Approach To Aspect Based Sentiment Analysis Using Transfer Learning}

\acrodef{ABSA}{Aspect-Based Sentiment Analysis}
\acrodef{MWEs}{multiword expressions}
\acrodef{LLM}{large language model}
\acrodef{ATE}{Aspect Term Extraction}
\acrodef{ASC}{Aspect Sentiment Classification}
\acrodef{NLP}{Natural Language Processing}
\name{
\begin{tabular}{c}
     Gaurav Negi\textsuperscript{1}, Rajdeep Sarkar\textsuperscript{2}$^{*}$\thanks{*Work done as a PhD student at University of Galway.}, Omnia Zayed\textsuperscript{1}, Paul Buitelaar\textsuperscript{1} 
\end{tabular}}

\address{\textsuperscript{1}Insight SFI Centre for Data Analytics, University Of Galway \\
          \textsuperscript{2}Yahoo Research \\
         \{gaurav.negi, omnia.zayed, paul.buitelaar\}@insight-centre.org\\
         rajdeep.sarkar@yahooinc.com\\
         }

\abstract{
\ac{ABSA} aims to identify terms or \ac{MWEs} on which sentiments are expressed and the sentiment polarities associated with them. The development of supervised models has been at the forefront of research in this area. However, training these models requires the availability of manually annotated datasets which is both expensive and time-consuming. Furthermore, the available annotated datasets are tailored to a specific domain, language, and text type. 
In this work, we address this notable challenge in current state-of-the-art ABSA research. We propose a hybrid approach for Aspect Based Sentiment Analysis using transfer learning. The approach focuses on generating weakly-supervised annotations by exploiting the strengths of both large language models (LLM) and traditional syntactic dependencies.  We utilise syntactic dependency structures of sentences to complement the annotations generated by LLMs, as they may overlook domain-specific aspect terms. Extensive experimentation on multiple datasets is performed to demonstrate the efficacy of our hybrid method for the tasks of aspect term extraction and aspect sentiment classification.
 \\ \newline  }

\begin{document}

\maketitleabstract
\section{Introduction}

\acf{ABSA}~\cite{liu2012absa} refers to the task of identifying the aspects of the entities and their associated sentiments from a given text sequence. \ac{ABSA} comprises two fundamental tasks: \ac{ATE} and \ac{ASC}. The significance of fine-grained \ac{ABSA} becomes apparent when different sentiments are articulated concerning distinct attributes of an entity, as highlighted by ~\citet{hu_mining_2004}. An Aspect Term is a single- or multi-word expression within the text that serves to describe a specific aspect or attribute of the entity upon which sentiment is being expressed. When applied to a collection of review sentences, \ac{ATE} involves the identification of all aspect terms or opinion targets contained within each sentence. Subsequently, \ac{ASC} is concerned with the classification of sentiments associated with each of the aspect terms that were identified during the ATE process.

For instance, consider the sentence: ``\textit{I liked the \underline{service} and the \underline{staff}, but not the \underline{food}.}'' This sentence conveys nuanced sentiments pertaining to specific aspects. In particular, the aspect terms \underline{service} and \underline{staff} are associated with a positive sentiment, while the aspect term \underline{food} carries a negative sentiment within the given context. It is crucial to emphasise the domain specificity of these aspect terms and the intricate relationships that exist between them as it enables a better understanding about the product, instead of directly analysing the sentiment of the text as a whole. In the example provided, the terms \underline{service}, \underline{staff}, and \underline{food} collectively suggest that the text originates from a restaurant review. A model trained for a different domain, such as ``Electronics'' would struggle to identify these domain-specific terms in the ``restaurant'' domain, thereby leading to sub-optimal performance in sentiment analysis tasks. This underscores the requirement for domain-specific \ac{ABSA} frameworks to enhance performance within specific domains. Nevertheless, constructing domain-specific datasets entails significant costs in terms of iterative efforts and the involvement of specialised personnel. Consequently, there arises a demand for the development of domain-specific \ac{ABSA} systems that can operate effectively without the necessity of manually annotated training datasets.

We initiate our investigation by employing syntactic dependencies to identify aspect terms within a domain-specific context.
However, our findings reveal that Machine Learning models trained on generic datasets are not universally applicable across various domains. Consequently, there arises a necessity for the domain adaptation of machine learning methodologies to address this task. Machine learning solutions generally treat \ac{ATE} and \ac{ASC} as supervised tasks. As supervised approaches rely on annotated datasets, their utility for \ac{ABSA} in diverse domains is inherently limited. In response to this limitation, unsupervised approaches have been explored to a limited extent in \ac{ABSA}, some of which utilise pseudo-labeling as an interim technique within their framework \cite{giannakopoulos-etal-2017-unsupervised, wu_hybrid_2018}.

Large Language Models (LLMs)~\cite{brown2020language} have ushered in a transformative era in the realm of \ac{NLP}. These models undergo pre-training on extensive unlabeled corpora and subsequent fine-tuning across various labeled downstream tasks. Research~\cite{wei2022finetuned, chung2022scaling} has demonstrated that fine-tuning over a multitude of tasks, each with specific instructions, enhances the capacity of a model to generalise to novel tasks that were not encountered during pre-training.

In our research, we leverage LLMs to address the \ac{ATE} task within a domain-specific context. Notably, we discover that while LLMs exhibit considerable power, they tend to exhibit limitations in capturing aspect terms with a higher recall rate, leading to suboptimal performance. This observation prompts a fundamental question: ``Can LLMs fine-tuned on a source domain, where abundant annotated training data is available, be effectively adapted to another target domain that lacks such annotated resources?''. To investigate this, we employ a transfer-learning approach, fine-tuning an LLM in the source domain (Gadgets and Social Media) and subsequently evaluating its performance on the target domain (Restaurant and Laptop). Our findings reveal that while the model adapts to address the in-domain ATE task, it still grapples with domain-specific challenges inherent to the target domains.

In response to these challenges, we introduce a hybrid solution designed to generate synthetic, domain-specific Aspect-Based Sentiment Analysis (ABSA) datasets through transfer learning. We leverages the LLM that has been fine-tuned for the ATE task, utilising it to generate domain-specific training data. This dataset is then enriched through the incorporation of syntactic dependencies, which enhances the recall of identified aspect terms. Subsequently, this refined training dataset is employed to conduct the fine-tuning of another LLM, resulting in the development of a domain-grounded ABSA model.
 In summary, our main contributions are as follows:
\begin{itemize}
    \item Method for domain adaptation with Transfer Learning for ATE, using LLMs. Including a hybrid method of annotating training datasets for ATE.
    \item Evaluating the performance of GPT-3.5-Turbo and Flan-T5 in zero-shot setting for the task of ATE and ASC. 
    \item Flan-T5 model fine-tuning and investigation of performance with qualitative and quantitative comparative analysis.
\end{itemize}

\section{Related Work}
The conception of ABSA entailed the feature-based summarisation of customer reviews of products sold online \cite{hu_mining_2004}. Features of the product on which the customers have expressed their opinions. Subsequently, a distinction was made between feature keywords and opinion words \cite{zhuang_movie_2006}. A holistic algorithm for feature-based sentiment analysis utilising context was introduced by \cite{ding2008liu}. Over time, this evolved into the research area of ABSA, where the tasks of ATE and ASC have been explored with different methods. 

\noindent \textbf{Rule Based Methods for ABSA}: These approaches utilised POS tagging, opinion lexicons \citelanguageresource{bing_lexicon} and the syntactic dependency parser to identify the syntactic relationship between opinion words and aspect terms. Almost all initial ABSA methods followed a rule-based approach \cite{hu_mining_2004, zhuang_movie_2006}. Double Propagation \cite{qiu_opinion_2011, qiu2009expanding} introduced methods for the dynamic expansion of opinion words. Automatic methods for effective dependency parsing rule selection were also introduced \citelanguageresource{liu_automated_2015}.

\noindent \textbf{Supervised Machine Learning Methods}: ATE can be formulated as a sequence labelling task and ASC as a classification task. Initial methods of note performed feature enrichment using a large unlabelled corpus (Amazon product reviews, Yelp Reviews) before using ML models like CRF \cite{wang_recursive_2016, toh_dlirec_2014} and SVM \cite{de-clercq-etal-2015-lt3}. Neural Network architectures that encode and utilise the sequential information with Long Short-Term Memory (LSTM) neural networks, Convolutional Neural Networks (CNN) and Attention were extensively used for ABSA. Some models also successfully utilised the relatedness of ATE and aspect-based sentiment analysis by multi-task learning with hard parameter sharing \cite{ruder17a_multi_task} in the deep learning model \cite{xue-etal-2017-mtna, liaspect}. The current state-of-the-art results are given by \cite{scaria_instructabsa_2023}, which utilises the LLM introduced by \cite{wang2022benchmarking}. It outperforms the previous state-of-the-art which was BERT-based models \cite{YangL2022}, \cite{zhang-etal-2022-towards}.

\noindent \textbf{Unsupervised Machine Learning Methods}: The objective of unsupervised ABSA methods is to avoid the usage of the annotated datasets. These approaches often are a combination of rule-based ABSA with heuristics to annotate datasets\cite{giannakopoulos-etal-2017-unsupervised, DBLP:journals/kbs/WuWWYH18, karaouglan2022extended}. 

Supervised machine learning approaches require domain-specific annotated datasets. Data annotation is expensive and time-consuming. Unsupervised methods are often complex and restrictive in their performance. Rule-based approaches are not as effective as the approaches that utilise contextual information effectively, i.e., LSTM, Transformers, BERT, T5. To the best of our knowledge  domain specificity and similarity of the sources of existing annotated datasets (reviews) exists in ABSA. In this paper, we describe an ABSA method based on transfer learning and domain adaptation. The approach focuses on generating weakly-supervised annotations by exploiting the strengths of both LLMs and traditional syntactic dependencies. We use LLMs because of their versatility, understanding of natural instruction and the generalised nature of learning through instruction fine-tuning \cite{brown2020language, wei2022finetuned}.
\section{Datasets}

\label{sec:dataset}
\begin{table}
  \centering
\resizebox{0.8\columnwidth}{!}{
\begin{tabular}{l|l|c|c}
 \toprule
 Source &Domain&Training &Testing\\
 \midrule
 SemEval (2014) & Laptop & 3041 & 800\\
 & Restaurant & 3045 & 800\\
\midrule
 Twitter Sentiments &Open Domain& 6248 & 692 \\
\midrule
 Gadget Reviews & Electronic& 2099 & - \\
 &Products &&\\
\bottomrule
\end{tabular}
}
\caption{\ac{ABSA} datasets}
\label{table:dataset_size}
\vspace{-5mm}
\end{table}
\vspace*{-\baselineskip}
We utilise the Restaurant and Laptop reviews dataset ~\citelanguageresource{pontiki-etal-2014-semeval}, Twitter Sentiments dataset~\citelanguageresource{dong2014adaptive} and the Garget reviews dataset~\citelanguageresource{liu_automated_2015} to conduct our experiments on different tasks.
The Restaurant and Laptop reviews dataset consists of reviews categorised into two categories: \textit{restaurant} and \textit{laptop}. The Gadget reviews dataset contains product reviews spanning three categories: \textit{speakers}, \textit{computers} and \textit{router}. Each review in the two datasets is annotated with aspect terms and their corresponding sentiment polarity. In the restaurant category, the sentences are also annotated with aspect category information and aspect category polarity. The Twitter Sentiments dataset provides manually annotated tweets for target-dependent sentiment analysis. Each tweet is annotated with a target term and its associated sentiment.
The size of the training and test split of the different datasets is detailed in Table~\ref{table:dataset_size}.

\section{Methodology}
\label{sec:methodology}
In this section, we begin with a formal definition of the problem statement. Subsequently, we delve into the process of creating domain-specific \ac{ATE} dataset utilising syntactic dependencies. We proceed to outline the experimental setup employed for assessing LLMs in the context of the ATE task  in a zero-shot learning scenario. Following this, we describe the configuration for domain transfer learning as applied to the Aspect-Based Sentiment Analysis (ABSA) task within LLMs. Finally, we describe our proposed methodology, \name, designed to facilitate the training of domain-specific ABSA models without necessitating the presence of manually annotated training data.



\subsection{Problem Formulation}
\label{sec:problem_formulation}
Given the input text $x = {x_1, x_2, ..., x_t}$, where $x_i$ represents the $i^{th}$ word in the input, \ac{ATE} pertains to the identification of single- or multi-word terms within $x$ that convey sentiments. Let $a_j = {x_k, x_{k+1}, ..., x_m}$ denote an aspect term, and \ac{ASC} involves the determination of the sentiment polarity $s_j$ associated with $a_j$. The sentiment $s$ can assume one of the values from the set {positive, negative, neutral}, denoted as $s \in \{positive, negative, neutral\}$.

\subsection{Syntactic Dependencies for ATE}
\label{sec:syntactic_dependencies}
Exploiting the syntactic dependency structures have been an integral part of the many ABSA approaches \cite{zhuang_movie_2006, qiu_opinion_2011, giannakopoulos-etal-2017-unsupervised}. The core concept underlying these approaches is the recognition of multiple syntactic relations that establish connections between opinion words and aspect terms. 
\begin{table}
\centering
\resizebox{0.8\columnwidth}{!}{
\begin{tabular}{l|c|c}
 \toprule
Dependency Relation & \thead{Aspect \\ Term(s)} & \thead{Opinion \\ Word}\\
 \midrule
 $AT \textrm{---}DEP\textrm{---}O$ & $AT$ & $O$ \\
 $AT\textrm{---}DEP_1\textrm{---}M\textrm{---}DEP_2\textrm{---}O$ & $AT$ & $O$ \\
 $AT_1\textrm{---}DEP_3\textrm{---}AT_2$ & $AT_1, AT_2$ & --- \\
 \bottomrule 
\end{tabular}
}
\caption{Syntactic Dependencies for ATE}
\label{table:dependecy_rules}
\vspace{-3mm}
\end{table}

We extract Noun Phrases (NP) from the text, considering them as potential aspect term candidates. To refine this candidate set, we perform a pruning step to eliminate all stop words and opinion words. Subsequently, we select the candidates that follow the syntactic dependency structures \citelanguageresource{dozat-etal-2017-stanfords} described in Table \ref{table:dependecy_rules} and designate them as aspect terms. $AT$, $AT_1$, and $AT_2$ are NPs, which were obtained by chunking. $O$ are the opinion words with the specific POS tags. The potential opinion words mentioned in the table $O$ are the words that are Adjectives($JJ$), verbs ($VB$) and adverbs ($RB$). $DEP$, $DEP_1$, $DEP_2$ and $DEP_3$ are the placeholders for universal dependency (UD) \cite{ud_2016} tags. They are used for defining syntactic relationships within the sentence. Their values are specified as follows: 

\begin{align}
&DEP \in (amod, nsubj, xcomp, obl, obj,nmod,dep) \nonumber\\
&DEP_1 \in (amod, nsubj, nmod) \nonumber\\
&DEP_2 \in (amod, nsubj, xcomp, advmod,nmod) \nonumber\\
&DEP_3 \in (conj)    \nonumber
\end{align}

$M$ is a placeholder for the words which are indirectly related to Aspect Term and Opinion which takes on the Noun ($NN$), Verb ($VB$) and Adverb ($RB$) POS tag values. For e.g. in the following sentence ``\textit{I liked the \underline{service} and the \underline{staff}, but not the \underline{food}.}'' $AT-DEP-O$ type relationship exists between \textit{\underline{service}} and \textit{liked}, $AT_1-DEP_3-AT_2$ type relationship between \textit{\underline{service}} and \textit{\underline{staff}} , enabling us to extract those two aspect terms.  

The syntactic dependency annotation method plays a crucial role in \ac{ABSA}. However, it is essential to acknowledge its limitations. Due to its lack of specificity, this method often exhibits lower precision than recall. Consequently, a notable issue arises: a significant number of terms are incorrectly identified as aspect terms. Addressing this challenge is vital for improving the accuracy of \ac{ATE}.

\subsection{Zero Shot ATE and ASC}
\label{sec:zero_shot}
In this section, we describe the zero-shot setup of LLMs for the \ac{ATE} and the \ac{ASC} task. In a zero-shot setting, we do not have access to labelled data, hence we leverage prompting \cite{prompt_2021b} to obtain aspect terms and their associated sentiments. Using the designated prompt templates, Prompt~\ref{prompt:ate} and Prompt~\ref{prompt:senti}, we input text instances denoted as $x_i$ into the LLM alongside the respective prompts. The template Prompt~\ref{prompt:ate} serves as a directive for the LLM, instructing it to discern the pertinent aspect terms within the given context. In tandem, Prompt~\ref{prompt:senti} functions to attribute a specific sentiment polarity to the AT identified in the preceding step. It is crucial to emphasise that this configuration operates in the absence of labeled data, and the LLM remains non-fine-tuned for either the \ac{ATE} or \ac{ASC} task.

\subsection{Transfer Learning for ATE and ASC}
\label{sec:transfer_learning}
In this configuration, we fine-tune a pre-trained Large Language Model (LLM) using the Twitter Sentiment and Gadget Reviews datasets, specifically for the ATE and ASC tasks. This fine-tuning phase is instrumental in facilitating the adaptation of the LLM to these tasks. Notably, the multi-domain characteristic of the Twitter Sentiment and Gadget Reviews datasets equips the LLM with the capability to generalise its performance across the ATE and ASC tasks. Subsequently, this fine-tuned model is harnessed within a transfer-learning framework for evaluation on domain-specific datasets.
In our approach, we engage in the fine-tuning of the LLM within a text-to-text format framework, wherein both the input and output are represented in textual form. Specifically, for both the ATE and ASC tasks, the LLM operates on input data in the form of prompt templates, yielding outputs in the form of either aspect terms or their corresponding sentiments. This fine-tuning process on multi-domain datasets serves the purpose of task adaptation, leveraging these prompt templates to enhance performance of the LLM.

\begin{tcolorbox}[left=0mm, top=0mm, right=0mm, toptitle=0mm, bottomtitle=0mm, boxsep=0.5mm]
    
\begin{prompt}
\footnotesize
\textbf{Input ($x_i$)} = Extract aspect terms from the following input. 

\textit{input: just watched ps i love you  on star movies . i love hilary swank's smile !} </s>

\textbf{Output ($y_i$)} = hilary swank </s>   
\captionsetup{singlelinecheck = false,  justification=raggedright}
\caption{\centering LLM Prompt for ATE}
\label{prompt:ate}
\end{prompt}

\end{tcolorbox}
\noindent \textbf{\acf{ATE}}: The model takes input $x_i$ and label $y_i$ at the training time. An example scenario is illustrated in Prompt~\ref{prompt:ate}.

\noindent \textbf{\acf{ASC}}: When predicting the sentiment class of an AT in $x_i$, we utilise the prmpt structure used in Prompt~\ref{prompt:senti}. The output label $y_i$ will be the polarity and can take values between (positive, negative or neutral). 
An example scenario is illustrated in Prompt~\ref{prompt:senti}.

\begin{tcolorbox}[left=0mm, top=0mm, right=0mm, bottomtitle=0mm, boxsep=0.5mm]
\begin{prompt}
\footnotesize
\textbf{Input ($x_i$)} = Given the aspect term and the sentence. Predict if the aspect term in the sentence has a positive, negative or neutral sentiment expressed on it. 

\textit{aspect term: hilary swank}

\textit{sentence: just watched ps i love you on star movies. i love hilary swank's smile!}</s>

\textbf{Output ($y_i$)} = positive</s>
\captionsetup{singlelinecheck = false,  justification=raggedright}
\caption{\centering LLM Prompt for ASC}
\label{prompt:senti}
\end{prompt}
\end{tcolorbox}
\subsection{Weakly Annotating Datasets for ATE}


Weakly annotated training datasets are used for adapting model to the training domain.In this section we discuss three methods of automatic annotations of the training text.

\subsubsection{Annotation With LLMs}
\begin{figure}[t] 
\centering
\adjustbox{width=6.8cm, height=5.5cm}{
\includegraphics{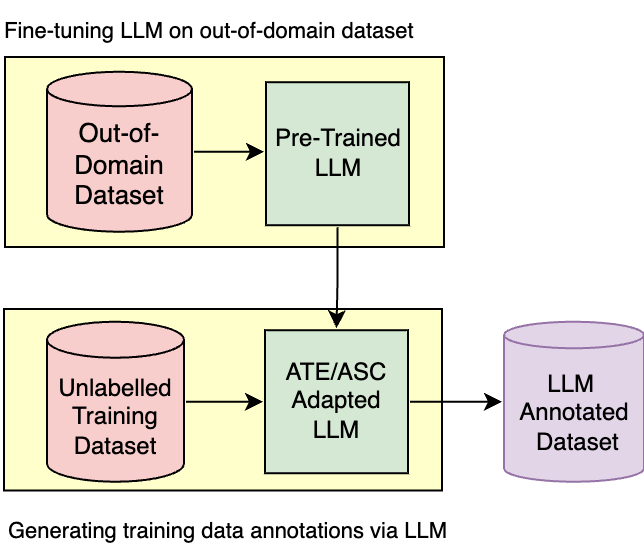}
}
\caption{Automatic annotations with LLM}
\label{fig:llm_annot}
\end{figure}
In our approach, we employ fine-tuned LLMs as described in Section~\ref{sec:transfer_learning} to generate domain-specific data annotations. Specifically, we utilise training data in templated forms, guided by prompts referenced as \ref{prompt:ate} and \ref{prompt:senti}, and generate annotations for each input text. This annotated training dataset based on LLMs serves as the foundation for training domain-adapted models. Notably, our hybrid annotation method also leverages this dataset.  

The entire process is graphically depicted in Figure \ref{fig:llm_annot}. Within this process, an out-of-domain dataset is utilised to fine-tune a pre-trained LLM  and this fine-tuning is carried out separately for the ATE and ASC task. The resulting out-of-domain adapted models are subsequently employed to annotate the in-domain training dataset, which, in turn, is used to fine-tune domain-specific ATE and ASC models.

\subsubsection{Hybrid Annotation Method}
\label{sec:hybrid_method}
\begin{figure*}[!ht] 
\centering{
\includegraphics[width=0.8\linewidth, height=2.5cm]{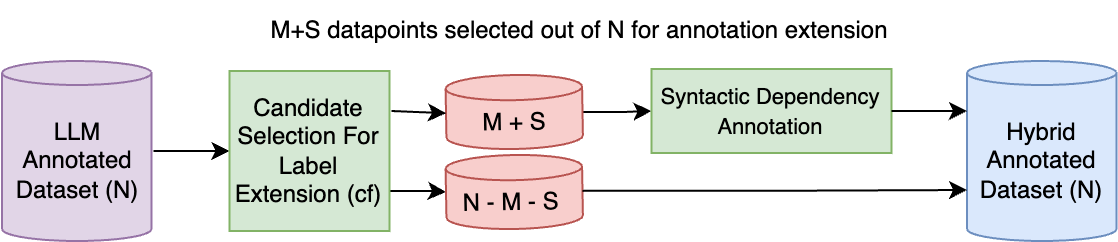}
}
\caption{Hybrid annotations}
\label{fig:hybrid_annot}
\end{figure*}


Aspect terms are known to exhibit a high degree of domain-dependence. Consequently, annotations generated by out-of-domain fine-tuned LLMs fail to fully grasp certain highly domain-specific aspect terms. To address this limitation, we strategically enhance LLM annotations by strategically incorporating the syntactic dependency annotation method.

We introduce a hybrid annotation approach that aims to systematically integrate the strengths of both methods. The syntactic dependency annotation method, known for its high recall, complements the LLM annotations, which are distinguished by their high precision. By carefully merging these two approaches, we aim to strike a balance that allows us to retain the precision advantages of the LLM-annotated training dataset while also benefiting from the broader coverage and recall offered by the syntactic dependency annotation method.

Figure \ref{fig:hybrid_annot} shows an overview of our hybrid annotation method for ATE. 
We begin with a dataset of unlabelled training text (N) with n sentences. The first step is to obtain high-precision annotations from a fine-tuned LLM (Flan-T5-Base-ATE) that is trained on a domain which is different from the domain of N. Once the dataset N has been labelled by the LLM, we consider a subset M of N, that contains at least one annotation by the LLM. From the remaining sentences in N that are not present in M, we select those sentences that have a high semantic similarity to sentences in M. The dataset of sentences thus selected be represented by S. Then we extend the annotations for (M + S) with the Syntactic dependency method.

\vspace{\baselineskip} \noindent \textbf{Candidate Selection for annotation extension}: We start with $n$ data points of LLM annotated dataset, where $N$ is their vector of texts from the dataset. We observe that the LLM annotation (Figure \ref{fig:llm_annot}) has a high precision and leaves out about 50\% of the training corpus without annotations. This situation encourages us to explore the integration of annotation methods. For this purpose, we consider two distinct categories of candidates from $N$ for label extension :
\begin{enumerate}
    \item The instances of $N$ that have at least one aspect term generated. This split of $m$ texts selected from the initial $n$ data points is represented as the vector $M$.
    \item The instances which have no annotations but are semantically similar to the annotated text ($s$) also from $n$ data points. This similar text split of size $s$ the data set with no annotation is represented as text vector $S$.
\end{enumerate}
We form these splits on the basis semantic similarity using sentence encoder \cite{sentence_enc_2019} to get $d$ dimensional vector representations of the sentences. We encode $M$ to vector $V_{M} \in \mathbb{R}^{m \times d}$. We calculate the mean vector $\mu_{M} \in \mathbb{R}^{1 \times d}$ from $V_M$ to select $s$ most similar text from $Q$ where $Q=N-M$ :
\begin{equation}
   V_{M} = \text{SentenceEncoder}(M) 
\end{equation}

\begin{equation}
   \mu_{M} =  \frac{\sum_{j=0}^{m} M_j}{m} 
\end{equation}

\noindent We select the $s$ most similar texts from $Q$ using the cosine similarity relative to $\mu_{M}$. To do so, we use the same sentence encoder we use with $M$ to get $V_M$ and encode $Q$ to $V_{Q}$.
\begin{equation}
   V_{Q} = \text{SentenceEncoder}(Q) 
\end{equation}

\noindent We then calculate the cosine similarity of all individual encoded text in $V_{Q}$, resulting in a similarity vector $Y$ which has 1:1 mapping with $V_{Q}$.
\begin{equation}
    \begin{split}
        Y = \text{CosineSim}(V_{Q}, \mu_{M})\\ 
        \text{where, } \text{CosineSim}: V_{Q} \mapsto Y
    \end{split}
\end{equation}
Cut-off fraction ($cf$) is the hyper-parameter that helps in calculating the cut-off value ($C_x$) for the similarity for a value from $Q$ to be included in $S$. This cut-off value is calculated in terms of mean ($\mu_{y}$) and standard deviation ($\sigma_{y}$) values of similarities $Y$.
\begin{equation}
    C_x = \mu_{y} + \sigma_{y} * cf
    \label{eq:cutoff}
\end{equation}

We then use $C_x$ as a threshold for similarity and select $S$ text vector from $Q$, which along with $M$ is passed through Syntactic Dependency annotation to extend the labels as shown in Figure \ref{fig:hybrid_annot}.
\vspace{-5mm}
\begin{equation}
X =
\begin{cases}
S & \text{, CosineSim}(V_{Q}, \mu_{M}) > C_x
       \\
    R & \text{, CosineSim}(V_{Q}, \mu_{M}) < C_x
       \\
\end{cases}
\label{eq2}
\end{equation}

After the cutoff filtering, we have $S$ with no aspect terms and $M$ texts with at least one aspect term. We combine $S+M$ and extract aspect terms with syntactic dependency structure as discussed in Section~\ref{sec:syntactic_dependencies}. If there are overlapping token in extraction aspect terms for $M$, we ignore the Dependency generated aspect terms in favour of LLM generated ones. At the end of this process we combine the newly annotated $(S, M)$ with $R$ non-annotated split to get Hybrid Annotated Dataset.

\section{Experiments}
\label{sec:experiments}
We start this section with the implementation details of the experiments. In Section~\ref{sec:results_discussion}, we perform a comparative analysis of the results.

\begin{table*}
  \centering
\resizebox{0.7\textwidth}{!}{
\begin{tabular}{l|c|ccc|ccc}
\toprule
&& \multicolumn{3}{c|}{Laptop Review 2014}&\multicolumn{3}{c}{Restaurant review 2014}\\
Method &Fine-Tuned&Precision&Recall&F1&Precision&Recall&F1\\
\midrule
Syntactic Dependency & No & 29.54&37.23&32.94&48.53&52.65&50.50\\
Flan-T5-Base (Zero Shot) & No & 12.71&17.07&14.57&19.83&15.09&17.14\\
Flan-T5-Base-ATE & Yes & 53.60&51.53&52.54&67.93&39.45&49.91\\
\bottomrule
\end{tabular}
}
\caption{ATE annotation efficiency on test splits}
\label{table:label_acc}
\end{table*}

\begin{table*}
  \centering
  \resizebox{0.8\textwidth}{!}{
\begin{tabular}{l|c|ccc|ccc}
\toprule
&& \multicolumn{3}{c|}{Laptop Review 2014}&\multicolumn{3}{c}{Restaurant review 2014}\\
Model & Type of Annotations &Precision&Recall&F1&Precision&Recall&F1\\
\midrule
GPT-3.5-TURBO & -- & 22.62&70.61&34.26&36.07&76.46&49.02\\
Flan-ATE-DOM-ADAPT & Flan-T5-ATE-Base & 66.07&45.84&54.13&77.95&43.68&55.99\\
Flan-ATE-DEP & Syntactic Dependency-based & 28.94&47.69&36.02&54.83&61.51&57.98\\
Flan-ATE-HYBRID & Hybrid & 47.85&60.15&53.50&55.57&69.54&61.77\\
\midrule
FLan-ATE-GOLD & Gold Labels &86.22&81.84&83.97&87.04&84.81&85.91 \\
\bottomrule
\end{tabular}
}
\caption{ATE efficiency after utilising training datasets with different annotation strategies}
\label{table:ate_acc}
\end{table*}

\subsection{Implementation Details}
\label{sec:implementation_details}
For our experimental setup, we utilise for the Flan-T5-Base\footnote{\tiny\url{https://huggingface.co/google/flan-t5-base}} language model as the LLM for our tasks. Adhering to the methodology outlined by ~\citet{chung2022scaling}, we employ Adafactor~\cite{shazeeradafactor} as our chosen optimiser and conduct training for 32 epochs using an NVIDIA Tesla V100 GPU. The best performing model out of 32 epochs is selected. The learning rate is set at 0.0003. Our training configuration involves a batch size of 4 with a gradient accumulation step size of 4, effectively simulating a batch size of 16. The sentence encoder utilised in the hybrid annotation method, as detailed in Section~\ref{sec:hybrid_method}, is the all-MiniLM-L6-v2\footnote{\tiny\url{https://huggingface.co/sentence-transformers/all-MiniLM-L6-v2}}. This encoder was selected from the sentence encoder leader-board. All hyper-parameters were determined through grid search.

\section{Results And Discussion}
\label{sec:results_discussion}
We compare our ATE/ASC approaches with zero-shot LLM predictions. To establish the upper bounds of performance metrics, we compare with the performance of a supervised model. Unlike our weakly supervised method, which uses automatically annotated training data, the supervised model relies on gold labels. The results of the ATE experiments are presented in Table 4 and 5. We evaluate ATE models in two steps:
\begin{enumerate}
    \item The first set of evaluations is testing our models and labelling methods on the test set directly. Here, we test the efficiency of our data generation method on test data without utilising training data for domain-adaptation. These results are presented in Table \ref{table:label_acc}. \textbf{Syntactic Dependency} method and \textbf{Flan-T5-Base (Zero Shot)} are the approaches which does not require any training while \textbf{Flan-T5-Base-ATE} uses Flan-T5-Base which has been fine-tuned on Gadget Reviews and Twitter sentiments.
    
    \item In the final step we evaluate the models that were fine-tuned on different annotations for the training dataset. These results are in Table \ref{table:ate_acc}.\textbf{Flan-ATE-GOLD} is fine-tuned on gold ATE labels. Training data annotated with Flan-T5-ATE-Base generation, using syntactic dependencies and hybrid method is used to fine-tune Flan-T5-Base to obtain \textbf{Flan-ATE-DOM-ADAPT}, \textbf{Flan-ATE-DEP} and \textbf{Flan-ATE-HYBRID} respectively.
\end{enumerate}

\begin{table*}
  \centering
  \resizebox{0.8\textwidth}{!}{
\begin{tabular}{l|c|ccc|ccc}
\toprule
&& \multicolumn{3}{c|}{Laptop Review 2014}&\multicolumn{3}{c}{Restaurant review 2014}\\
Model & Type of Annotations &Precision&Recall&F1&Precision&Recall&F1\\
\midrule
GPT-3.5-TURBO & -- & 74.42&75.69&74.95&71.53&74.47&72.87\\
Flan-T5-BASE & None (Zero Shot) & 63.89&67.87&63.67&63.09&66.97&63.54\\
Flan-T5-TL & (Twitter Sentiments + gadget review) &74.65&70.04&68.74&70.76&70.15&70.16\\
Flan-T5-DOM-ADAPT & (Flan-T5-Base-InDomain-Gen) &80.76&79.81&79.55&74.35&75.19&74.72\\
\midrule
FLan-T5-GOLD & Gold Labels &78.71&80.47&78.91&84.73&80.20&82.23 \\
\bottomrule
\end{tabular}
}
\caption{ASC results}
\label{table:senti_acc}
\end{table*}
The results pertaining to Aspect-Based Sentiment Classification (ASC) are presented in Table \ref{table:senti_acc}. All models utilise Prompt \ref{prompt:senti} for predictions and fine-tuning. The evaluation metrics are macro-averaged across sentiment classes, with any aspects exhibiting conflicting predictions removed from the test dataset. Each prediction entails using the text and aspect terms from the test data to predict sentiments. We perform sentiment predictions without fine-tuning on both GPT-3.5-Turbo and Flan-T5-Base for zero-shot inference. Flan-T5-TL is Flan-T5-Base fine-tuned on Twitter Sentiments and gadget data. Flan-T5-DOM-ADAPT is trained on in-domain generated polarities using Flan-T5-TL, while Flan-T5-GOLD is trained on the original gold sentiment labels of the training data.

\subsection{Quantitative Evaluation}

\noindent \textbf{ATE Results}: For the zero-shot evaluation as seen in Table \ref{table:label_acc}, Flan-T5-Base (Zero Shot) has considerably low accuracy scores, meaning that it does a poor job of annotating the test data for aspect terms. The score syntactic dependency method of annotation is higher than zero-shot. We see that the model (Flan-T5-ATE-Base), fine-tuned on the Gadget Reviews and Twitter Sentiments, shows an 
improvement in the annotation generation capabilities.

The results presented in Table \ref{table:ate_acc} pertain to models that underwent fine-tuning using training data annotated through various annotation methods, signifying the domain adaptation process for \ac{ATE}. The upper performance bound is established by a model that was fine-tuned using gold ATE labels in a purely supervised fashion. Notably, the specifics of the data utilised for training GPT-3.5-Turbo are not disclosed, which is evident in the observed results. GPT-3.5-Turbo exhibits notably high recall but lower precision, resulting in an F1-score that reflects its tendency to extract a multitude of aspect term candidates with relative inaccuracy. Models trained with syntactic dependency annotation achieve comparatively favorable performance in only one of the domains, albeit still exhibiting suboptimal precision values. Conversely, when we train a model with annotations generated by Flan-T5-ATE-Base, a model adapted for the task using out-of-domain data, we observe a more balanced performance across both domains.

Incorporating the hybrid annotation method, our primary objective is to amalgamate annotation techniques characterised by high precision (Flan-T5-ATE-Base) and high recall (syntactic dependency annotations) in order to enhance the overall F1 score of the fine-tuned model (Flan-ATE-HYBRID). It is evident that, in the restaurant domain, this approach surpasses the F1-score in comparison to the previously discussed methods. Conversely, in the case of Laptop Reviews, when the model is trained using hybrid annotation for the training dataset, we observe superior performance in terms of F1-score compared to the model exclusively trained on syntactic dependency annotations and GPT-3.5-TURBO. Nevertheless, the F1-score remains slightly lower than that achieved by the ATE model trained on annotations generated by Flan-T5-ATE-Base.

\noindent \textbf{ASC results}: The results pertaining to ASC in Table \ref{table:senti_acc} reveal a notable zero-shot performance of the Flan-T5-BASE model. We posit that this zero-shot performance can be attributed to the efficiency of Flan fine-tuning tasks, as demonstrated by Chung et al.~\cite{chung2022scaling}. Flan-T5-TL, fine-tuned on out-of-domain data, exhibits improvements over the zero-shot performance. GPT-3.5-Turbo, a model of considerable scale and fine-tuning tasks, outperforms Flan-T5-TL, possibly due to its larger size and diverse fine-tuning objectives. Moreover, we observe that the domain-adapted model (Flan-T5-DOM-ADAPT), which underwent fine-tuning on in-domain auto-annotated training data using Flan-T5-TL, outperforms GPT-3.5-TURBO. Furthermore, the model Flan-T5-GOLD, fine-tuned on gold annotated training data, demonstrates robust evaluation metrics. Making it a good upper bound reference for the performance of our ASC models.

\begin{table*}[t]
  \centering
  \resizebox{0.8\textwidth}{!}{
\begin{tabular}{l|c|c|c|c|c|}
\toprule
Text & GPT-3.5-TURBO & Flan-ATE-DOM-ADAPT & Flan-ATE-HYBRID & GOLD \\
\toprule
Two wasted steaks -- what a crime! & steaks, wasted, crime & - & steaks & steaks \\
\midrule
You will be very happy with the experience. & experience, happiness & experience & experience & - \\
\midrule
We didn't know if we should order a drink or leave? & drink, leave & - & drink & drink \\
\midrule
\multicolumn{1}{c|}{The room is a little plain, but it's difficult to make such a}& room, plain,& room & room, place & room,\\
small place exciting and I would not suggest that as a  & small size,& & & place\\
reason not to go. & excitement& & & \\

\bottomrule
\end{tabular}
}
\caption{ATE Prediction examples}
\label{table:ate_eg}
\end{table*}
\subsection{Qualitative Evaluation}
\subsubsection{ASC Results}
\begin{figure}[t] 
\centering
\adjustbox{width=6cm, height=5cm}{
\includegraphics{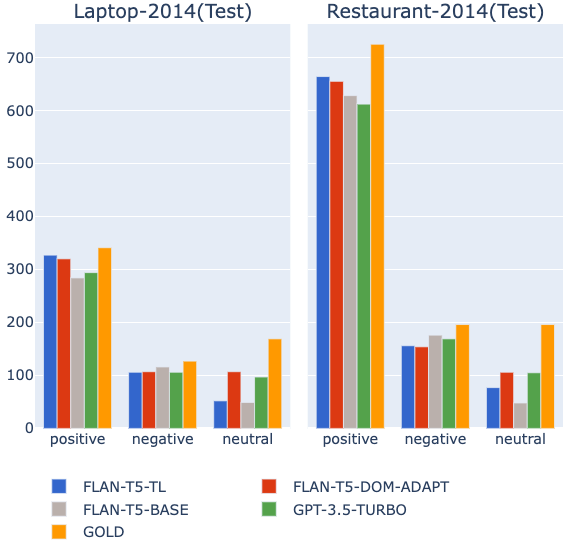}
}
\caption{Class-wise Analysis Of ASC models}
\label{fig:at_senti_qual}
\end{figure}

In Figure \ref{fig:at_senti_qual}, we present a comparative analysis of class-wise correct predictions made by models using the gold labels from the test set across both domains. Flan-T5-TL and Flan-T5-DOM-ADAPT represent our fine-tuned models, the details of which have been previously elucidated. Notably, the models exhibit a certain level of difficulty in accurately predicting the \textit{neutral} labels. GPT-3.5-Turbo and the domain-adapted Flan-T5-DOM-ADAPT excel in correctly predicting the \textit{neutral} class. This difficulty in predicting the \textit{neutral} class has an impact on the macro-averaged metrics. However, when predicting \textit{positive} and \textit{negative} classes, Flan-T5-TL, Flan-T5-DOM-ADAPT, and Flan-T5-Base (zero-shot) showcase comparable, if not superior, performance compared to GPT-3.5-TURBO.

In Table \ref{table:ate_eg}, we present some key results pertaining to ATE. GPT-3.5-TURBO exhibits a high number of false positive aspect terms, which can be attributed to its significantly higher recall but lower precision. Conversely, Flan-ATE-DOM-ADAPT demonstrates high precision, which leads to missed aspect terms. In contrast, Flan-ATE-HYBRID strikes a balance between precision and recall, resulting in occasional false extractions but fewer compared to GPT-3.5-TURBO. This balance contributes to its higher F1 score.


\subsubsection{Impact of CF (cutoff fraction) on Performance}

\begin{figure}[t]
\centering
\vspace{-2mm}
\begin{tabular}{cc}
\includegraphics[width=0.45\columnwidth]{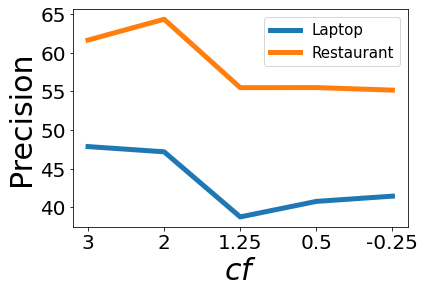} &
\includegraphics[width=0.45\columnwidth]{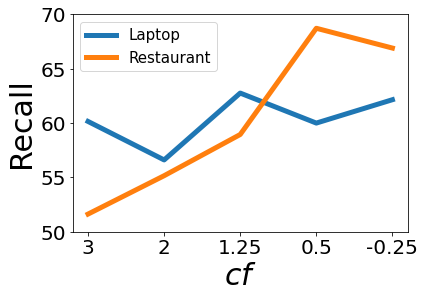}
\end{tabular}
\vspace{-3mm}
\caption{Impact of $cf$ on ATE Hybrid Models}
\label{fig:cf_effect}
\vspace{-3mm}
\end{figure}

In Section~\ref{sec:hybrid_method}, Equation~\ref{eq:cutoff} incorporates the cutoff fraction ($cf$) to determine the similarity cut-off ($C_x$) for the hybrid annotation method. A higher value of $cf$ corresponds to an elevated similarity cut-off, resulting in a more stringent criterion for data points to be labeled using syntactic dependency as the value of $C_x$ decreases.

In Figure \ref{fig:cf_effect}, we illustrate the effective control of the precision-recall trade-off achieved through the manipulation of the parameter $cf$, which governs the annotation splits and contributes to more efficient learning. Lowering the value of $cf$ results in a reduction of the cut-off similarity, thereby allowing a greater number of data points with diminishing similarity to undergo label extension via syntactic dependency structures. Notably, we observe an ascending trend in recall values as $cf$ is decreased. It is worth highlighting an intriguing trend in precision: initially, there is a sharp decline, but with a well-balanced inclusion of segments and exclusion of segments in the annotation process, we witness a subsequent increase in precision. This suggests the potential for further improvements in model performance when operating with larger unannotated datasets.

\vspace{-2mm}
\section{Conclusion}
\vspace{-2mm}
In our work, we undertake a systematic analysis of the performance of non-learnable syntactic dependencies and the application of transfer learning in Large Language Models (LLMs) for the Aspect-Term Extraction (ATE) task in Aspect-Based Sentiment Analysis (ABSA). Subsequently, we introduce an innovative hybrid framework for generating domain-specific datasets by leveraging task-adapted LLMs and syntactic dependencies for both ATE and Aspect-Based Sentiment Classification (ASC) tasks in ABSA. We execute comprehensive experiments across multiple datasets to illustrate the robust empirical performance of our proposed approach compared to other baseline methodologies. In terms of future work, we envisage extending this research to encompass multilingual settings and investigating the utilization of encoder-only architectures.

\section{Limitations}

We utilise syntactic dependencies and LLMs for generating synthetic dataset. However, LLMs and syntactic dependencies have limited capabilities in a low-resource setting. In the future, we intend to address this limitation. We would also like to extend the scope of our experiments to more open-sourced LLMs and explore the utilisation of encoder-only architecture too.


\section{Acknowledgements}
This work was conducted with the financial support of the Science Foundation Ireland (SFI) under Grant Number SFI/12/RC/2289\_P2 (Insight\_2) and was also supported by funding from the Irish Research Council (IRC) for the Postdoctoral Fellowship award GOIPD/2023/1556.



\section{Bibliographical References}
\bibliographystyle{lrec-coling2024-natbib}
\bibliography{lrec-coling2024-example}

\section{Language Resource References}
\bibliographystylelanguageresource{lrec-coling2024-natbib}
\bibliographylanguageresource{languageresource}

\nocite{*}



\end{document}